\documentclass[tbtags]{article}

\usepackage{mlspconf}
\usepackage{amssymb,amsfonts}
\usepackage{graphicx}
\usepackage{booktabs}
\usepackage{tabularx}
\usepackage{float}
\usepackage{placeins}
\usepackage{xcolor}
\usepackage{algorithm}
\usepackage{algpseudocode}
\usepackage[font=footnotesize,labelfont=bf]{caption}
\usepackage{mathtools}   
\usepackage{adjustbox}
\usepackage{hyperref}
\copyrightnotice{979-8-3503-2411-2/25/\$31.00 {\copyright}2025 IEEE}

\toappear{2025 IEEE International Workshop on Machine Learning for Signal Processing, Aug.\ 31-- Sep.\ 3, 2025, Istanbul, Turkey}

\title{COORDINATE ASCENT NEURAL KALMAN-MLE FOR STATE ESTIMATION}
\name{%
    Bettina Hanlon$^{\star}$%
    \qquad Ángel F. García-Fernández$^{\dagger}$\thanks{This work was supported by the EPSRC Centre for Doctoral Training in Distributed Algorithms EP/S023445/1 and the Government Communications Headquarters (GCHQ).}%
}
\address{%
    $^{\star}$ Department of Electrical Engineering and Electronics, University of Liverpool, UK \\%
    $^{\dagger}$ IPTC, ETSI de Telecomunicaci\'on, Universidad Polit\'ecnica de Madrid, Spain%
}

\begin{document}

\maketitle

\begin{abstract}
This paper presents a coordinate ascent algorithm to learn dynamic and measurement models in dynamic state estimation using maximum likelihood estimation in a supervised manner. In particular, the dynamic and measurement models are assumed to be Gaussian and the algorithm learns the neural network parameters that model the dynamic and measurement functions, and also the noise covariance matrices. The trained dynamic and measurement models are then used with a non-linear Kalman filter algorithm to estimate the state during the testing phase.
\end{abstract}

\section{Introduction}
\label{sec:intro}

Bayesian filtering algorithms have a wide array of applications, including positioning and tracking \cite{positioning1}. These filtering algorithms perform state estimation of a dynamic system that is observed indirectly through a series of noisy measurements \cite{cupbookchapter1}. Examples of different Bayesian filters include the Kalman filter, the extended Kalman filter \cite{EKF}, the unscented Kalman filter (UKF) \cite{unscentedKF}, the iterated extended Kalman filter \cite{Bell93} and the iterated posterior linearisation filter \cite{E}.

Filtering algorithms rely on knowledge of the state-space model, including the dynamic and measurement models. Sometimes, these models are not known, so its parameters must be learnt. To do so, one option is to use parameter estimation techniques \cite{cupbookchapter1,Gao19}, for instance, using Markov chain Monte Carlo or expectation maximisation \cite{Kokkala16}.

An alternative option is to use a more flexible approach to learn the dynamic and measurement models (or parts of them), for instance using Gaussian processes \cite{6099561, NIPS2013_2dffbc47} and neural networks \cite{Imbiriba2024AugmentedPM, GEDON2021481}, and then to use a Bayesian filtering algorithm based on the learnt models. Examples of approaches that fuse these two methods include the work \cite{uuv} which uses a convolutional neural network particle filter to deal with the problem of unmanned underwater vehicle target state estimation. Reference \cite{dynanet} presents a hybrid deep learning and time-varying state space model which can be trained end to end with examples in visual odometry, sensor fusion and motion prediction. Another approach in \cite{etoe} utilises a neural network to learn the proposal distribution of a particle filter by approximating it as a multivariate Gaussian mixture and learning the relevant means and covariances. A recent approach, KalmanNet, assumes knowledge of the dynamic and measurement functions with unknown noise covariance matrices and aims to learn the Kalman gain \cite{Kalmannet}. KalmanNet is based on a training set with sequences of states and measurements (supervised learning). 

In a supervised learning scenario, it is also possible to perform maximum likelihood estimation (MLE) of the parameters in the training phase and Bayesian filtering in the testing phase \cite{betti2024}. For the same problem formulation as KalmanNet (known functions, but unknown noise covariance matrices), this approach, called Kalman-MLE approach, has been shown to outperform KalmanNet in both accuracy and training time in \cite{betti2024}.

The contribution of this paper is to extend Kalman-MLE to scenarios in which the measurement and dynamic functions are also unknown. That is, we deal with supervised learning and develop an MLE algorithm to estimate the dynamic and measurement functions, and the covariance matrices of the process and measurement noises, making the problem formulation more general than in \cite{Kalmannet, betti2024}. In particular, we model the dynamic and measurement functions as neural networks and the coordinate ascent MLE algorithm iteratively learns the neural network weights and the noise covariances. During testing phase, a non-linear Kalman filter \cite{cupbookchapter1} is run with the learned models. The resulting algorithm is referred to as coordinate ascent neural Kalman-MLE (CAN-Kalman-MLE). Numerical experiments show the benefits of CAN-Kalman-MLE compared with KalmanNet.


\section{Problem Formulation and Background}
\label{Problem Formulation and Background}
This section presents the models and reviews parameter estimation using MLE in supervised learning.
\subsection{Models and data}

This paper considers the problem of Bayesian filtering \cite{cupbookchapter1} where the dynamic and measurement functions and their noise covariances are unknown. We denote the target state at time step $k$ as $x_k\in\mathbb{R}^{n_{x}}$. The corresponding noisy measurement through which it is observed is denoted by $z_k\in\mathbb{R}^{n_{z}}$. The way in which the target state evolves from state $x_{k-1}$ to $x_k$, dependent on the unknown parameter $\theta_g$, is represented by the transition density $g_{\theta_{g}}\left(x_{k}|x_{k-1}\right)$. Analogously, given the target state and dependent on the unknown parameter $\theta_l$, the measurement density is given by $l_{\theta_{l}}\left(z_{k}|x_{k}\right)$.

As in KalmanNet \cite{Kalmannet}, for the training phase, we are provided with $M$ sample sequences of ground truth states with their corresponding measurements, indexed by $i$, and denoted by $\left(x_{0:T}^{i},z_{1:T}^{i}\right)$ where $x_{0:T}^{i}=\left(x_{0}^{i},...,x_{T}^{i}\right)$ and $z_{1:T}^{i}=\left(z_{1}^{i},...,z_{T}^{i}\right)$ and $T$ is the sequence length.

\subsection{Parameter estimation via MLE}
Given the $M$ sample sequences of states and measurements, which are assumed independent, their joint density is \cite{cupbookchapter1}
\begingroup
\begin{multline}
p_{\theta_{g},\theta_{l}}\left(x_{0:T}^{1:M},z_{1:T}^{1:M}\right)  \\ =\prod_{i=1}^{M}\left[p\left(x_{0}^{i}\right)\prod_{k=1}^{T}g_{\theta_{g}}\left(x_{k}^{i}|x_{k-1}^{i}\right)\prod_{k=1}^{T}l_{\theta_{l}}\left(z_{k}^{i}|x_{k}^{i}\right)\right].
\label{5}
\end{multline}
\endgroup

The MLE of the parameters is then \cite{betti2024}
\begingroup
\begin{equation}
\hat{\theta}_{g} =\underset{\theta_{g}}{\mathrm{argmax}}\sum_{i=1}^{M}\sum_{k=1}^{T}\ln g_{\theta_{g}}\left(x_{k}^{i}|x_{k-1}^{i}\right),
\label{9}
\end{equation}
\endgroup
\begingroup
\begin{equation}
\hat{\theta}_{l} =\underset{\theta_{l}}{\mathrm{argmax}}\sum_{i=1}^{M}\sum_{k=1}^{T}\ln l_{\theta_{l}}\left(z_{k}^{i}|x_{k}^{i}\right).
\label{10}
\end{equation}
\endgroup
Once we have learned $(\theta_g,\theta_l)$, we can apply a Bayesian filtering algorithm to estimate the state when new data is available.

\section{Coordinate ascent MLE training}
\label{Coordinate ascent MLE training}
This section presents the dynamic and measurement models and the proposed CAN-Kalman-MLE training to learn the parameters of the dynamic and measurement functions and the noise covariance matrices.

\subsection{Dynamic and measurement models}
We consider additive Gaussian models such that the transition and measurement densities are \cite{cupbookchapter1}: 
\begin{equation}
\label{dyn12}
g_{\theta_{g}}\left(x_{k}|x_{k-1}\right) = \mathcal{N}\left(x_{k};f_{\theta_{g}}\left(x_{k-1}\right),Q\right)\\
\end{equation}
\begin{equation}
l_{\theta_{l}}\left(z_{k}|x_{k}\right) = \mathcal{N}\left(z_{k};h_{\theta_{l}}\left(x_{k}\right),R\right)
\end{equation}
where $f_{\theta_g}(\cdot)$ is the dynamic function,  $h_{\theta_l}(\cdot)$ is the measurement function, $Q$ is the process noise covariance matrix, and $R$ is the measurement noise covariance matrix.

We model functions $f_{\theta_g}(\cdot)$ and $h_{\theta_l}(\cdot)$ as neural networks with known architectures, and the neural network weights are $\theta_g$ and $\theta_l$. The parameters to learn also include the covariance matrices $Q$ and $R$ \cite{cupbookchapter1}. 

We propose a coordinate ascent algorithm \cite{Nocedal_book99} to iteratively learn the neural network weights and the noise covariance matrix via MLE, see \eqref{9} and \eqref{10}. Once the trained models have been obtained, these are then integrated in a non-linear Kalman filter, such as the UKF, in the testing phase. For details of non-linear Kalman filtering, we refer the reader for instance to \cite{cupbookchapter1}.

\subsection{Coordinate ascent MLE training}

In this section we present the coordinate ascent MLE algorithm to learn the dynamic model. The algorithm is directly applicable to the measurement model, which can be learnt in parallel by making the appropriate changes. From (\ref{9}) and (\ref{dyn12}), the estimated parameters and the estimated covariance matrix of the process noise are \cite{betti2024}:
\begin{align}
\label{coorddes}
\left(\hat{\theta}_{g},\hat{Q}\right) 
& = \underset{\theta_{g},Q}{\arg\min} \Bigg[
    MT\ln\left(|Q|\right) \notag \\
& \hspace{-1.1cm}
    + \sum_{i=1}^{M}\sum_{k=1}^{T}
    \left(x_{k}^{i}-f_{\theta_{g}}\left(x_{k-1}^{i}\right)\right)^{T}
    Q^{-1}
    \left(x_{k}^{i}-f_{\theta_{g}}\left(x_{k-1}^{i}\right)\right)
\Bigg].
\end{align}
It is in principle possible to optimise \eqref{coorddes} directly by defining a custom loss function in a neural network framework. However, this direct approach would not take into account that $Q$ is a covariance matrix, so special care must be taken in its optimisation. This paper proposes a coordinate ascent MLE approach (coordinate descent of (\ref{coorddes})) in which we iteratively optimise over $\theta_g$ and $Q$.

Let \( \hat{Q}_{j} \) denote the estimated value of \( Q \) at the $j$-th iteration.  
For this value of \( \hat{Q}_{j} \), we can estimate the dynamic function parameter \( \hat{\theta}_{g,j} \) as:
\begingroup
\begin{align}
\label{eighteen}
\hat{\theta}_{g,j} 
& = \underset{\theta_{g}}{\arg\min} \Bigg[
    MT\ln\left(|\hat{Q}_{j}|\right) 
    + \sum_{i=1}^{M}\sum_{k=1}^{T}
    \left(x_{k}^{i}-f_{\theta_{g}}\left(x_{k-1}^{i}\right)\right)^{T} \notag \\
& \quad \times \hat{Q}_{j}^{-1}
    \left(x_{k}^{i}-f_{\theta_{g}}\left(x_{k-1}^{i}\right)\right)
\Bigg] \notag \\
& = \underset{\theta_{g}}{\arg\min} 
    \sum_{i=1}^{M}\sum_{k=1}^{T}
    \left(x_{k}^{i}-f_{\theta_{g}}\left(x_{k-1}^{i}\right)\right)^{T}
     \notag \\
& \quad \times \hat{Q}_{j}^{-1} \left(x_{k}^{i}-f_{\theta_{g}}\left(x_{k-1}^{i}\right)\right).
\end{align}
\endgroup
It should be noted that in (\ref{eighteen}) we are optimising over neural network weights, for which we can use neural network optimisation frameworks, which often make use of stochastic gradient descent (SGD) methods. In the experimental results, we perform the optimisation using the ADAM optimiser.

For the given \( \hat{\theta}_{g,j} \), the next estimated value of the noise covariance matrix is
\begin{align}
\label{nineteen}
\hat{Q}_{j+1} 
& = \underset{Q}{\arg\min} \Bigg[
    MT\ln\left(|Q|\right) \\
&   \hspace{-1cm} + \sum_{i=1}^{M}\sum_{k=1}^{T}
    \left(x_{k}^{i}-f_{\hat{\theta}_{g,j}}\left(x_{k-1}^{i}\right)\right)^{T} \notag Q^{-1}
    \left(x_{k}^{i}-f_{\hat{\theta}_{g,j}}\left(x_{k-1}^{i}\right)\right)
\Bigg] \notag \\
& \hspace{-1cm} = \frac{1}{MT} \sum_{i=1}^{M} \sum_{k=1}^{T}
    \left(x_{k}^{i}-f_{\hat{\theta}_{g,j}}\left(x_{k-1}^{i}\right)\right) \notag\left(x_{k}^{i}-f_{\hat{\theta}_{g,j}}\left(x_{k-1}^{i}\right)\right)^{T}.
\end{align}
This optimisation admits a closed-form solution \cite{pml1Book}, which is convenient for implementation. Then, the coordinate ascent MLE algorithm performs steps (\ref{eighteen}) and (\ref{nineteen}) iteratively until convergence, or until a specified number of iterations has taken place. The pseudocode is provided in Algorithm 1.
\begin{algorithm}[htb]
\setlength{\textfloatsep}{0pt}
\caption{Coordinate Ascent MLE for Learning the Dynamic Model}
\label{alg:coord_ascent_MLE}
\begin{algorithmic}[1]
\State \textbf{Model Inputs:} 
\State \quad Training sequences $\{x_k^i\}$ for $i=1,\dots,M$, $k=0,\dots,T$.
\State \quad Learning rate $\alpha$, batch size $|B|$, number of epochs $N_e$.
\State \quad Number of coordinate ascent iterations, $N_c$.
\State \textbf{Output:} Optimized parameters $\hat\theta_{g\ast}$, $\hat{Q}_\ast$.
\vspace{1mm} 

\State \textbf{1.} Initialize neural network weights $\hat\theta_{g,0}$ randomly.
\State \textbf{2.} Initialize $\hat{Q}_{0}$, for instance, as an identity matrix.
\State \textbf{3.} \textbf{For} $j = 1$ to $N_c$:
\State \quad \textbf{(a) Update $\hat\theta_{g,j}$ with fixed $\hat{Q}_{j-1}$:}
\State \quad \quad \textbf{For} $epoch = 1$ to $N_e$:
\State \quad \quad \quad \textbf{For} each mini-batch  $B\subseteq\left\{ 1,...,M\right\} \times\left\{ 1,...,T\right\} $ of data $\left(x_{k-1}^{i},x_{k}^{i}\right)$:
\State \quad \quad \quad \quad Compute residuals $r_k^i = x_k^i - f_{\hat\theta_{g,j}}(x_{k-1}^i)$.
\State \quad \quad \quad \quad Compute mini-batch loss (\ref{eighteen}) for $(i,k)\in B$:
\[
\mathcal{L}(\theta_g) = \sum_{(i,k)\in B} (r_k^i)^T \hat{Q}_{j-1}^{-1} r_k^i
\]
\State \quad \quad \quad \quad Update $\theta_{g}$ to yield $\hat{\theta}_{g,j}$.
\State \quad \quad \quad \textbf{End for (mini-batches)}
\State \quad \quad \textbf{End for (epochs)}
\State \quad \textbf(b) Update $\hat{Q}_j$ using the current $\hat\theta_{g,j}$ in (\ref{nineteen}).
\State \quad \textbf{End for (outer cycles $j$)}
\end{algorithmic}
\end{algorithm}

\section{Experimental results}
\label{results}
This section presents the experimental results. We test the proposed CAN-Kalman-MLE algorithm\footnote{Python implementation will be made available at https://github.com/berthanlon/CAN-Kalman-MLE.} in two scenarios. In particular, we test the estimation capabilities when learning different combinations of known/unknown and dynamic/measurement functions and their noise covariance matrices. When the dynamic and measurement functions are known, there is no coordinate descent algorithm in CAN-Kalman-MLE, and the algorithm reduces to Kalman-MLE \cite{betti2024}. We compare the results with KalmanNet, which knows the dynamic and measurement function but does not know the noise covariance matrices, and with a UKF \cite{unscentedKF}, which knows the dynamic and measurement model.

\subsection{Dataset and algorithm parameters}
In all scenarios, synthetic trajectories and measurements were sampled from the dynamic and measurement models described  in their respective scenario section. In all cases, $1200$ simulated trajectories were generated. These were split into $M=1000$ training sequences, and $200$ testing sequences. Various combinations of noise parameter magnitudes and sequence lengths $T$ were tested. During training, we divide the $M\times T$ data points into mini-batches by reshaping them into a 2D format.

The architecture of the neural networks that model the dynamic and measurement functions is shown in Figure 1. It was determined through experimentation with the number of layers and the size of each hidden layer. The hidden layer size, dropout rate and training hyper-parameters were manually tuned based on validation performance. Optimal combinations of these vary between scenarios. 

In these experiments, for the dynamic model, we actually learn the function that maps the difference with the previous time step. That is, we assume that $f(x_{k-1})=x_{k-1}+f_{\hat{\theta}_{g,j}}(x_{k-1})$. In this way, when we run Algorithm 1, the input to the neural network $f_{\hat{\theta}_{g,j}}$ is $x_{k-1}$ and the target is $\Delta{x_k}$, where $\Delta{x_k} = x_k - x_{k-1}$. For the measurement model, we have the input $x_k$ and target $z_k$. 

\begin{figure}
    \centering
    \hspace*{0cm}
    \includegraphics[scale=0.23]{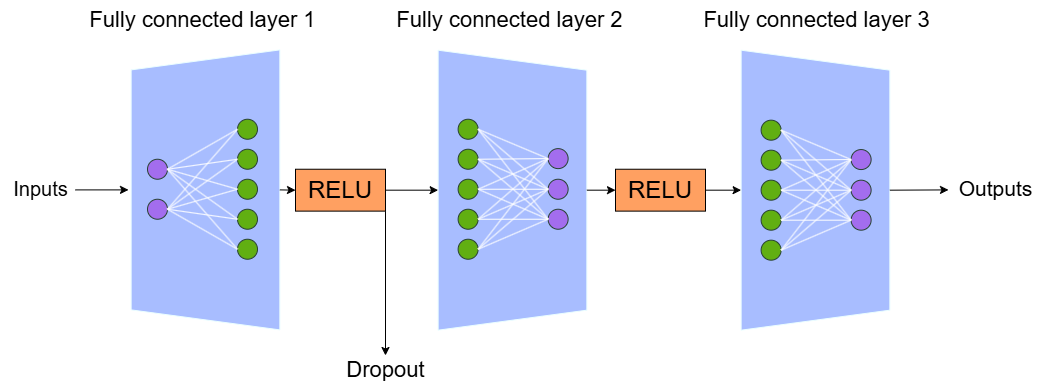}
    \caption{{Block diagram showing the architecture of the multi-layer perceptron used to learn the dynamic function $f(\cdot)$. It consists of three fully connected layers, two ReLU activation functions and a dropout layer after the first ReLU. Exact inputs and outputs vary dependent on scenario.}}
    \label{blocks}
\end{figure}

Testing with these learned models is done with a UKF. The UKF has been implemented with the sigma points in \cite{unscentedKF}, with parameters $\alpha = 0.1$, $\beta = 3$, and $\kappa = 0$. KalmanNet was trained with 400 epochs, a batch size of 30, learning rate of $1\times10^{-3}$ and weight decay of $1\times10^{-5}$ for both scenarios.


\subsection{Scenario 1: Bilateration-based target tracking}
\label{sc2}

In this scenario, a state $x_k = [p^x_k, v^x_k, p^y_k, v^y_k]^T$  includes the positions in $x$ and $y$, and their respective velocities. The state transition function and its respective covariance are given by a nearly constant velocity model \cite{Bar-Shalom_book01,betti2024} with sampling time $\tau=0.5$ and process noise parameter $\sigma_u$, which is varied in the experiments. The prior at time step $0$ is Gaussian with mean $\bar{x}_0 = [100,1,0,2]^T$ and $P_0 = \mathrm{diag}([1,0.1,1,0.1])$. 
\begin{figure}
 \hspace*{-0.5cm} 
 \includegraphics[scale=0.4]{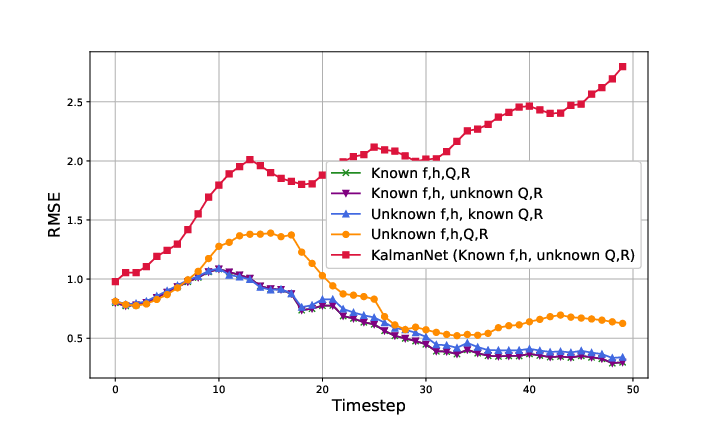}
 \caption{RMSE at every time step for Scenario 1 where $T=50$ and $\sigma_u^2 = 0.001$ and $\sigma_r^2 =0.001$.}
\end{figure}

The measurements contain the range obtained from two radars, positioned at (0,0) and (150,0): 
\begin{align}
z_k = 
\underbrace{\begin{pmatrix}
\sqrt{{x_{1,k}}^2 + {x_{3,k}}^2} \\
\sqrt{({x_{1,k}}-150)^2 + {x_{3,k}}^2})
\end{pmatrix}}_{\text{$h$}(x_k)}
+
\underbrace{\begin{pmatrix}
v_{1,k} \\
v_{2,k}
\end{pmatrix}}_{\text{$v$}_k}
\end{align}
where the covariance matrix of the noise is $R=\sigma_r^2 I_2$. The initial guess for the noise covariances to begin training are $\hat{Q}_0 = I_4 $ and $\hat{R}_0 = I_2$.

\begin{table}[h]\label{hyperparamtable}
    \centering
    \setlength{\tabcolsep}{3pt} 
    \renewcommand{\arraystretch}{0.9} 
    \caption{Hyper-parameters for learning the functions in Scenario 1.}
    \label{tab:model_hyperparams}
    \footnotesize 
    \begin{tabular}{|l|l|c|c|c|c|c|}
        \hline
        \textbf{Function} & \textbf{Noise Cov.} & \textbf{Hidden L. Size} & \textbf{Learning Rate} & \textbf{$N_c$} & \textbf{$N_E$} & \textbf{$\lvert B \rvert$} \\
        \hline
        $h$ & K  & 256  & $1 \times 10^{-3}$ & -  & 600  & 160 \\
        $h$ & U  & 256  & $1 \times 10^{-3}$ & 10 & 800  & 160 \\
        $f$ & K  & 600  & $1 \times 10^{-4}$ & -  & 150  & 32  \\
        $f$ & U  & 600  & $1 \times 10^{-4}$ & 10 & 150  & 32  \\
        \hline
    \end{tabular}
\end{table}

\begin{table*}[h!]
\centering
\caption{Offline training and online inference times (in seconds) for CAN-Kalman-MLE for different configurations of known and unknown \( f, h, Q, R \) across 25-step and 50-step simulations in Scenario 1.}
\label{TrainingInferenceTimes}
\scriptsize
\setlength{\tabcolsep}{8pt}
\begin{tabular}{l | cc | cc | cc | cc | cc}
\toprule
\textbf{$T$} & \multicolumn{2}{c|}{\textbf{Known \( f, h, Q, R \)}} 
             & \multicolumn{2}{c|}{\textbf{Known \( f, h \), Unknown \( Q, R \)}}
             & \multicolumn{2}{c|}{\textbf{Unknown \( f, h \), Known \( Q, R \)}}
             & \multicolumn{2}{c|}{\textbf{Unknown \( f, h, Q, R \)}}
             & \multicolumn{2}{c}{\textbf{KalmanNet}} \\  
\cmidrule(lr){2-3} \cmidrule(lr){4-5} \cmidrule(lr){6-7} \cmidrule(lr){8-9} \cmidrule(lr){10-11}
& \textbf{Training} & \textbf{Testing} 
& \textbf{Training} & \textbf{Testing} 
& \textbf{Training} & \textbf{Testing} 
& \textbf{Training} & \textbf{Testing} 
& \textbf{Training} & \textbf{Testing} \\
\midrule
25   & -  & 00:00:00 &  00:00:02 & 00:00:00 &  00:32:47 &  00:00:08 & 00:44:11  & 00:00:11 & 00:22:30 & 00:00:04 \\
50   & - & 00:00:01 & 00:00:04 & 00:00:01 &  01:06:16 & 00:00:33 & 01:28:21  & 00:00:45  & 01:01:59 & 00:00:09 \\
\bottomrule
\end{tabular}
\end{table*}

\begin{figure}
 \hspace*{-0.5cm} 
 \includegraphics[scale=0.4]{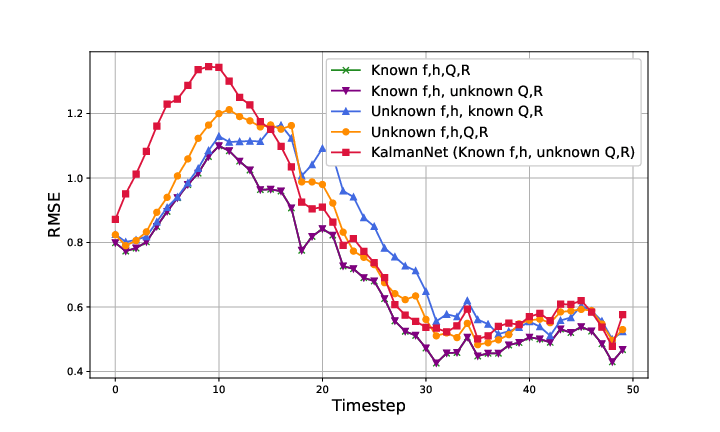}
 \caption{\normalsize{RMSE at every time step for Scenario 1 where $T=50$ and $\sigma_u^2 = 0.1$ and $\sigma_r^2 =1$.}}
\end{figure}

Table 1 shows the hyperparameters (including hidden layer size, learning rate, and coordinate ascent parameters) used to learn the unknown dynamic and measurement functions, $f$ and $h$, in the two cases where their corresponding noise covariance matrices are either known or unknown (K and U in the table). 

Figures 2 and 3 show two cases in which there are differences between the methods. We can see that the UKF (known parameters) and Kalman-MLE (known $f$, $h$ and unknown $Q$ and $R$) offer similar performance. KalmanNet has considerably worse performance, especially in Figure 2, even though it has the same information ($f$ and $h$) and the same training and testing data. It is interesting that CAN-Kalman-MLE with unknown models outperforms KalmanNet in these cases. This is likely due to the fact that CAN-Kalman-MLE has been derived from first mathematical principles (MLE). 

Table 2 shows the computational time, (hours: minutes: seconds) required for offline training and online inference for each method. These times have been obtained running the algorithms in a desktop with a 3.60 GHz AMD Ryzen 5 5500 and a GeForce-RTX4070 graphics card. Kalman-MLE (known $f$, $h$) is the fastest algorithm to train. KalmanNet has a longer training time, and CAN-Kalman-MLE is the most computationally expensive method, probably due to the nested 'for' loops required for the coordinate ascent algorithm. Nevertheless, CAN-Kalman-MLE also considers more unknown parameters (functions and covariance matrices) than KalmanNet and Kalman-MLE (only unknown covariance matrices).

\noindent\subsection{Scenario 2: Lorenz attractor}
\begin{table}[h]
    \centering
    \hspace{-1cm}
    \setlength{\tabcolsep}{3pt} 
    \captionsetup[table]{font=footnotesize,labelfont=bf}
    \renewcommand{\arraystretch}{0.9} 
    \caption{Hyper-parameters for learning the functions in Scenario 2.}
    \label{tab:model_hyperparams}
    \footnotesize 
    \begin{tabular}{|l|l|c|c|c|c|c|}
        \hline
        \textbf{Function} & \textbf{Noise Cov.} & \textbf{Hidden L. Size} & \textbf{Learning Rate} & \textbf{$N_c$} & \textbf{$N_E$} & \textbf{$\lvert B \rvert$} \\
        \hline
        $h$ & K  & 400  & $1 \times 10^{-3}$ & -  & 1500 & 32 \\
        $h$ & U & 400  & $1 \times 10^{-3}$ & 40  & 3000 & 32 \\
        $f$ & K & 400  & $1 \times 10^{-3}$ & -  & 300 & 32  \\
        $f$ & U & 400  & $1 \times 10^{-3}$ & 50 & 2000 & 32  \\
        \hline
    \end{tabular}
\end{table}

\begin{table*}[h!]
\centering
\caption{RMSE across all time steps for Scenario 2 for various combinations of known and unknown \( F, H, Q, R \) and KalmanNet.}
\label{RMSEtable_FHQR_KNet}
\scriptsize
\setlength{\tabcolsep}{6pt}
\begin{tabular}{l | c | c c c c c}
\toprule
\textbf{$T$} & \( r \) & \textbf{Known \( F, H, Q, R \)} & \textbf{Known \( F, H \), Unknown \( Q, R \)} & \textbf{Unknown \( F, H \), Known \( Q, R \)} & \textbf{Unknown \( F, H, Q, R \)} & \textbf{KalmanNet} \\ 
\midrule
     & \( 10^{-5} \)  &   0.0254 & 0.0151 &  1.1558 & 2.7603 &  0.2188 \\
     & \( 10^{-4} \)  & 0.04605     &0.04398 & 2.0505 & 2.3042  &  0.3526 \\
  25 & \( 10^{-3} \)  &  0.1378& 0.1433  &  0.7088 &  1.1859 &  0.4791  \\
     & \( 10^{-2} \)  &  0.3738  &  0.3867   &   0.7090 &  0.9341 &  1.0077      \\
\hline
     & \( 10^{-5} \)  &    0.0288 &  0.0115 &  0.8383 &  1.2337 &  0.0035 \\
     & \( 10^{-4} \)  &    0.0410 &     0.0341 & 1.1525 &  1.7344 &  0.2789\\
 50  & \( 10^{-3} \)  &   0.0919  & 0.0963  & 1.9992 & 3.2513  & 6.1358 \\
     & \( 10^{-2} \)  &     0.2895 &  0.3038 & 0.9554 &  0.6226 & 0.4099  \\
\bottomrule
\end{tabular}
\end{table*}

In this scenario, we consider that the dynamic model is a discretised Lorenz attractor \cite{Kalmannet}. The state is $x_k = [x_{k,1}, x_{k,2}, x_{k,3}]^T$ and the resulting discrete-time evolution process, with a 3-D state, is given by:
\begin{equation}
    {x}_{k+1} = {f}({x}_{k}) + {w}_k = {F}({x}_{k}) \cdot {x}_{k} + {w}_k
\end{equation}
where $w_k$ is a zero-mean Gaussian noise with covariance matrix $Q$ and
\begin{equation}
F(x_k) \triangleq \exp\left(A(x_k) \cdot \Delta k\right) \approx I_3 + \sum_{j=1}^{J} \frac{\left(A(x_k) \cdot \Delta k\right)^j}{j!} 
\end{equation}
where \(J\) represents the number of coefficients.
\begin{equation}
\label{lor1}
A(x_\tau) =
\begin{pmatrix}
-10 & 10 & 0 \\
28 & -1 & -x_{k,1} \\
0 & x_{k,1} & -\frac{8}{3}
\end{pmatrix}.
\end{equation}
The prior mean at time step $0$ is $\bar{x}_0 = [1,1,1]^T$ and covariance matrix is $P_0 = \gamma^2 I_3$ where $\gamma^2 = 1\times 10^{-2}$. 

The measurement model is defined as:
\begin{equation}
    z_k = h_{\text{spherical}}(x_k) + v_k,
\end{equation}
where:
\begin{equation}
    h_{\text{spherical}}({x_k}) = \sqrt{x_{k,1}^2 + x_{k,2}^2  + x_{k,3}^2},
\end{equation}
which provides the radial distance of the state vector from the origin in the 3-D state space.

The noise covariance matrices are \cite{Kalmannet}
\begin{equation}
Q = q^2 \cdot I_3, \quad R = r^2 , \quad \nu \triangleq \frac{q^2}{r^2}.
\end{equation}
We set $\nu=0.01$. As a result, $r$ was varied for experiments, along with $T$, resulting in the variation in $Q$. The initial guess for the noise covariances into the neural network to begin training were $\hat{Q}_0 = I_3 $ and $\hat{R}_0 = 1$. Table 3 shows the hyper-parameters used to train the unknown models in the cases where their noises are known or unknown.

Table 4 shows RMSE across all time steps in each situation. CAN-Kalman-MLE performs similarly to the situation where noise parameters are known, and only the functions are modelled. The UKF is generally the best performing filter as it knows the dynamic and measurement models exactly. This is followed by Kalman-MLE, with known functions but unknown covariance matrices. KalmanNet usually performs worse than these two methods. CAN-Kalman-MLE with unknown functions performs worse than the UKF and Kalman-MLE with known functions, as expected.

\section{Conclusions}\label{sec: conclusion}
We have proposed CAN-Kalman-MLE, a supervised learning, coordinate ascent MLE algorithm that iteratively learns the neural network parameters that model dynamic and measurement functions, and the covariance matrices of the noises. It has been demonstrated that in some cases it performs close to a UKF where the dynamic and measurement models are known. Also, CAN-Kalman-MLE can outperform KalmanNet, which is significant as KalmanNet has access to the dynamic and measurement functions. This is probably due to the fact that CAN-Kalman-MLE has been derived based on a mathematically principled approach via MLE.

A line of future work is to design neural Kalman-filter-type algorithms in an unsupervised case, where there is no access to ground truth states and corresponding measurements for training, as for instance in \cite{Imbiriba2024AugmentedPM}.

%

\bibliographystyle{IEEEbib}
\bibliography{bibliography}

\end{document}